\documentclass[letterpaper]{article}
\usepackage{aaai}
\usepackage{times}
\usepackage{helvet}
\usepackage{courier}

\usepackage{amsmath}
\usepackage{amssymb}
\usepackage{amsfonts}
\usepackage{multirow}
\usepackage{graphicx} 
\usepackage{tabularx} 
\usepackage{url}
\usepackage{orcidlink}
\providecommand{\href}[2]{#2}
\usepackage{comment}
\usepackage{booktabs}
\usepackage{tcolorbox}
\usepackage{xcolor}

\tcbuselibrary{skins,breakable}
\definecolor{promptbg}{RGB}{240,248,255} 
\definecolor{promptborder}{RGB}{100,149,237} 

\newtcolorbox{promptbox}[1][]{
  colback=promptbg,
  colframe=promptborder,
  fonttitle=\bfseries,
  boxrule=0.5pt,
  arc=3pt,
  breakable,
  enhanced jigsaw,
  left=3pt,
  right=3pt,
  top=3pt,
  bottom=3pt,
  #1
}
 
\newif\ifanonymous
\anonymousfalse  

\ifanonymous
\else
\fi
\begin{document}
%
\title{SAGE: Hierarchical LLM-Based Literary Evaluation\\through Ontology-Grounded Interpretive Dimensions}

\author{Tianyu Wang\,\orcidlink{0000-0002-9244-8798}\\
twang4@mercy.edu\\
Mercy University, Math \& Computer Science Department\\
Dobbs Ferry, NY, USA
\And
Nianjun Zhou\,\orcidlink{0000-0002-3473-6097}\\
jzhou@us.ibm.com\\
IBM T.J. Watson Research Center\\
Yorktown Heights, NY, USA}
\nocopyright
\maketitle

\begin{abstract}

Evaluating literary quality requires assessing interpretive dimensions such as cultural representation, emotional depth, and philosophical sophistication that resist straightforward computational measurement. We introduce SAGE, a hierarchical evaluation framework that decomposes literary quality into ontology-grounded interpretive dimensions assessed through structured large language model evaluation with multi-round iterative reflection and independent validation.

We validate the framework on 100 short stories (50 canonical works, 30 pulp fiction, 20 LLM-generated narratives) across three analytical layers (cultural, emotional-psychological, existential-philosophical) using dual-mode assessment. Across 600 evaluations, the framework achieves 98.8\% score convergence and greater than 94\% inter-rater agreement, with near-perfect mode invariance between content-based and metadata-based evaluation. Statistical analysis reveals a consistent genre hierarchy (Canonical $>$ Pulp $>$ LLM, all $p<0.001$) with layer-specific discrimination: cultural critique and philosophical depth exhibit very large effect sizes (Cohen's $d>2.4$), while emotional representation shows smaller gaps ($d$=1.68), suggesting that affective patterns are more learnable from training data than critical stance or philosophical depth. Cross-layer correlations ($r$=0.649--0.683) confirm the three dimensions capture empirically distinguishable quality facets. These findings demonstrate that theory-driven LLM evaluation can achieve measurement-grade reliability and support systematic identification of where current generative models fall short of human literary production, with direct implications for scalable automated evaluation of open-ended text generation.

\end{abstract}

\section{Introduction}\label{sec:introduction}

Recent advances in large language models (LLMs) have enabled machines to generate increasingly sophisticated narrative texts, including short stories, scripts, and even full-length novels. These developments raise a fundamental question for both artificial intelligence and the humanities: can machine-generated narratives achieve the kinds of literary qualities traditionally associated with human-authored works? While substantial research has focused on improving the generative capabilities of language models, comparatively little attention has been devoted to the problem of evaluating the literary quality of the narratives they produce.

To illustrate the challenge of literary evaluation, consider the opening sentence of Franz Kafka’s \emph{The Metamorphosis}~\cite{kafka1915metamorphosis}:

\begin{quote}
``When Gregor Samsa woke one morning from uneasy dreams, he found himself transformed in his bed into a gigantic insect.''
\end{quote}

This sentence establishes several elements commonly associated with significant literary works: a distinctive narrative voice, a striking symbolic premise, and an atmosphere that invites psychological and philosophical interpretation. Over the course of the story, Gregor Samsa’s transformation has been interpreted as representing alienation, existential anxiety, and the fragile relationship between the individual and society. These meanings emerge not merely from the surface structure of the sentence but from deeper interactions among narrative context, thematic development, and reader interpretation.

However, most automatic evaluation metrics used in natural language processing would fail to capture these aspects. Measures such as BLEU, ROUGE, and BERTScore \cite{papineni2002bleu,lin2004rouge,zhang2019bertscore} were designed primarily for tasks such as machine translation or summarization and rely on lexical or embedding similarity to reference texts. While useful in those contexts, such metrics correlate only weakly with human judgments in open-ended generation tasks and are particularly ill-suited for evaluating narrative fiction \cite{novikova2017we,reiter2018structured}. Literary works are not defined solely by surface similarity to existing texts; rather, they are characterized by complex interactions among narrative structure, thematic meaning, emotional resonance, cultural context, and philosophical interpretation. While computational methods can reliably assess Kafka's linguistic precision, syntactic structure, and thematic organization---properties corresponding to lower analytical layers (L1--L3)---they cannot evaluate the symbolic, psychological, and existential dimensions that define the story's canonical significance.

This challenge reflects a deeper conceptual issue. Literary evaluation does not correspond to a single, theory-neutral notion of quality that can be reduced to a scalar score. Instead, literary criticism traditionally examines works across multiple interpretive dimensions. Some aspects of evaluation concern observable textual and structural properties that can be analyzed through linguistic and narrative methods, such as grammatical fluency, discourse coherence, and thematic organization. These aspects have been widely studied in computational linguistics and computational narratology. Prior research has developed methods for analyzing textual cohesion and coherence \cite{graesser2004coh}, modeling discourse structure and entity-based coherence \cite{barzilay2008modeling}, and identifying narrative patterns in texts \cite{elson2010automatic,chambers2008unsupervised}. Such approaches suggest that lower-level narrative properties can often be analyzed with reasonable reliability using automated methods.

In contrast, other aspects of literary evaluation involve interpretive constructs that emerge from the interaction between text, reader, and cultural context. These include cultural representation, emotional--psychological experience, and philosophical reflection. These dimensions of literary interpretation have traditionally been the domain of human readers and critics and are difficult to capture using conventional text similarity metrics or surface-level linguistic analysis. Recent advances in large language models suggest that such models may be capable of reasoning about higher-level narrative meaning, making them promising tools for structured literary evaluation.

In this work, we propose \textbf{SAGE} (Systematic Assessment of Generative Excellence), a hierarchical evaluation framework organized around \emph{ontological decomposition}: the recognition that evaluative dimensions differ in epistemic kind and therefore require distinct assessment methods. Throughout this paper, \emph{ontology} refers to a structured taxonomy of evaluative dimensions by epistemic type, not to formal knowledge representations or semantic-web ontologies. The lower layers focus on observable textual and narrative properties that can be analyzed using established NLP techniques, while the higher layers capture interpretive dimensions of literature, including cultural representation, emotional--psychological experience, and existential or philosophical engagement. By structuring evaluation across these layers, the framework provides a multidimensional approach to assessing narrative texts rather than reducing literary quality to a single numerical score. In practice, our primary focus is on the higher layers (L4--L6), since the lower layers (L1--L3) can be effectively handled by existing rule-based and statistical methods, whereas the interpretive dimensions in L4--L6 require the reasoning and contextual understanding capabilities that have only recently become available through modern large language models.

The purpose of this framework is not to improve the generation of AI-written narratives, but to examine whether machine-generated stories can satisfy the same dimensions of literary evaluation traditionally applied to human-authored works. In particular, we investigate whether narratives produced by large language models exhibit properties corresponding to multiple layers of literary interpretation. By focusing on the higher interpretive layers (L4--L6), we address a fundamental question in computational literary studies: can automated systems engage with the cultural, emotional, and philosophical dimensions that have traditionally required human critical judgment?

To explore these questions empirically, we examine narratives drawn from three broad categories: canonical literary works, commercial genre fiction, and stories generated by large language models. These categories are not treated as absolute indicators of literary quality; rather, they represent different contexts in which literary evaluation historically occurs. We introduce two complementary evaluation settings. In \emph{content-limit evaluation}, assessment is based directly on the textual content of a narrative work. In \emph{title-limit evaluation}, assessment relies on external critical knowledge associated with a work's title or reputation, approximating forms of evaluation that occur in broader literary discourse. Comparing these two settings allows us to investigate how different sources of information influence literary judgments.

Our study is guided by the following research questions:

\textbf{RQ1 (Reliability):} Can LLM-based literary evaluation produce stable and reproducible assessments across multiple evaluative layers?

\textbf{RQ2 (Discriminative Capacity):} Do evaluations differentiate meaningfully among narratives drawn from different contexts, including canonical literature, genre fiction, and LLM-generated texts?

\textbf{RQ3 (Evaluation Robustness):} How do evaluation outcomes change under different informational conditions, such as direct textual analysis versus reputation-based contextual knowledge?

\textbf{RQ4 (Dimensional Independence):} Are the proposed evaluative dimensions non-redundant at two levels: within each layer (e.g., do affective complexity and psychological interiority within Layer 5 capture distinct properties of emotional representation?) and across layers (do cultural, emotional-psychological, and existential-philosophical representation constitute genuinely separate quality dimensions?), supporting the view that literary quality is inherently multidimensional?

\paragraph{Contributions.}
This work makes three contributions. First, we introduce \textbf{SAGE} (Systematic Assessment of Generative Excellence), a hierarchical framework that decomposes literary quality into six analytically distinct layers, separating rule-based assessment of observable textual properties (L1--L3) from LLM-based evaluation of interpretive dimensions (L4--L6). Each interpretive dimension is grounded through exemplar contrasts drawn from the literary canon, demonstrating that the twelve dimensions are non-redundant: canonical works occupy divergent positions on theoretically motivated axes (e.g., high affective complexity with near-zero psychological interiority in Hemingway versus both high in Woolf), establishing construct validity independently of experimental results. Second, we design a dual-track evaluation architecture comprising five-round iterative self-reflection and independent cross-validation, with layer-specific bias mitigation strategies that enable reliable, reproducible assessment of interpretive constructs at scale. Third, we present a controlled empirical study comparing canonical literature, commercial genre fiction, and LLM-generated narratives across three interpretive layers, revealing a capability profile that distinguishes pattern-reproducible literary skills from stance-requiring ones and identifies where current generative models fall systematically short of human literary achievement. Code, evaluation prompts, dataset, and results are publicly available at \url{https://github.com/tisage/sage-eval}.

\section{Related Work}
\label{sec:related}

Standard NLG metrics including BLEU (n-gram precision against reference translations~\cite{papineni2002bleu}), ROUGE (recall-oriented overlap designed for summarization~\cite{lin2004rouge}), and perplexity measure surface similarity to reference texts and correlate poorly with human judgments of narrative quality \cite{reiter2018structured,novikova2017we}. Embedding-based approaches like BERTScore \cite{zhang2019bertscore} improve semantic coverage through contextual representations, yet still struggle with discourse-level properties spanning extended narratives. Applied to Kafka's \emph{Metamorphosis}, such metrics might capture lexical patterns or sentence-level coherence but would miss the symbolic weight of transformation as alienation, the psychological complexity of Gregor's interiority, or the existential critique of modern life, precisely the interpretive dimensions that define literary significance.

Narrative-specific frameworks have extended evaluation to textual and structural quality dimensions. The HANNA benchmark \cite{chhun2022human} provides 1,056 human-annotated stories evaluated across six criteria including coherence, empathy, and engagement, with systematic comparison revealing that existing metrics capture different aspects of quality with limited agreement. NarraBench \cite{hamilton2025narrabench} presents a theory-informed taxonomy of narrative understanding tasks, surveying 78 existing benchmarks to systematize evaluation approaches. Yang and Jin \cite{yang2024makes} identify accuracy, completeness, appropriateness, insight, and consistency as frequently assessed quality dimensions, while noting the absence of unified frameworks capturing these dimensions simultaneously. The SCORE framework \cite{yi2025score} evaluates character consistency, emotional coherence, and plot tracking through dynamic state tracking. Complementary approaches have examined discourse coherence through entity distribution patterns \cite{barzilay2008modeling}, emotional trajectory analysis \cite{reagan2016emotional}, and fractal sentiment structures \cite{bizzoni2023fractality}. Together, these approaches capture surface-level and structural properties of narrative texts but do not address the interpretive dimensions that distinguish canonical literature from competent but unremarkable prose.

Cultural representation, emotional-psychological sophistication, and philosophical engagement, the interpretive dimensions central to literary criticism, have received limited attention in computational evaluation frameworks, despite substantial work on related but distinct problems.

\textbf{Cultural Representation.} NLP research has extensively studied bias detection and fairness in language models~\cite{blodgett2020language,sap2020social}, examining how texts encode stereotypes, power asymmetries, and cultural assumptions. However, this work focuses primarily on identifying harmful representations rather than evaluating the sophistication of cultural engagement. Literary evaluation requires assessing whether texts demonstrate nuanced understanding of cultural systems, power structures, and social relations, questions informed by cultural theory (Bourdieu's cultural capital, Said's Orientalism, Spivak's subaltern studies) rather than fairness metrics. No existing framework systematically evaluates cultural representation as a dimension of literary quality.

\textbf{Emotional-Psychological Depth.} Sentiment analysis and emotion recognition systems classify texts along valence dimensions or basic emotion categories~\cite{poria2019emotion}. While useful for understanding affective patterns, these approaches do not capture the complexity, ambiguity, and granularity that characterize sophisticated emotional representation in literature. Literary evaluation requires assessing affective complexity (contradictory emotions, ambiguity), psychological interiority (access to characters' inner worlds), and emotional granularity (precise differentiation of feeling states), dimensions grounded in affect theory and narrative psychology rather than sentiment classification. Existing computational approaches do not evaluate emotional-psychological sophistication as a literary quality.

\textbf{Philosophical Engagement.} Topic modeling~\cite{blei2003latent} and theme extraction identify recurring concepts and semantic patterns in texts. However, identifying that a text discusses mortality or freedom differs fundamentally from evaluating the depth of its philosophical engagement with these themes. Literary evaluation requires assessing whether texts grapple substantively with existential questions (authenticity, meaning, finitude), moral complexity (ethical ambiguity, competing frameworks), and the human condition (vulnerability, mortality, meaning-making), dimensions informed by existential philosophy, moral philosophy, and philosophical anthropology. No computational framework evaluates philosophical depth as a dimension of literary quality.

LLM-based evaluation has emerged as a scalable alternative to human annotation. \cite{zheng2023judging} demonstrate that strong LLMs achieve over 80\% agreement with human preferences on MT-Bench and Chatbot Arena, establishing viability for automated evaluation while identifying systematic biases including position bias, verbosity bias, and self-enhancement bias. \cite{li2024llms} provide a comprehensive survey noting that LLM judges correlate more highly with non-expert human judges than expert annotators, raising questions about whether such correlation reflects deep quality understanding or surface-level pattern matching. Most LLM-as-judge work focuses on conversational quality, instruction following, or general text generation rather than literary or narrative evaluation. The few studies addressing creative writing employ single-pass assessment without explicit theoretical grounding, leaving open whether structured frameworks can improve reliability and interpretive depth.

AI-generated narratives now span long-form generation systems facing systematic coherence degradation \cite{wu2025longeval}, hierarchical planning approaches \cite{gu2025rapid,wan2025cognitive}, multi-agent narrative architectures \cite{huot2024agents,ran2025bookworld}, and collaborative authoring tools \cite{yuan2022wordcraft,mirowski2023co}, together producing increasingly sophisticated texts that demand evaluation capable of distinguishing nuanced quality differences. As LLMs generate narratives approaching human-level surface fluency, the question shifts from whether machines can produce grammatical stories to whether they achieve the interpretive depth characterizing canonical literature---a question that existing evaluation paradigms, designed for surface-level and structural properties, are not equipped to answer.

The gaps identified in this review, namely the absence of ontology-grounded evaluation for interpretive dimensions (L4--L6), the lack of structured multi-round assessment for literary analysis, and the need for systematic comparison across authorship types, directly motivate the framework and methodology described in the following section.

\section{Methodology}
\label{sec:methodology}

\subsection{Framework Overview}

We propose a six-layer hierarchical framework that decomposes literary quality into increasingly interpretive dimensions. The framework distinguishes between lower layers (L1-L3) that assess objective textual properties through rule-based computational metrics and upper layers (L4-L6) that evaluate interpretive literary qualities through LLM-based critical analysis. This decomposition reflects an empirically testable hypothesis: formal textual features and interpretive literary qualities constitute largely independent dimensions requiring different analytical approaches.

\begin{figure*}[t]
\centering
\includegraphics[width=\textwidth]{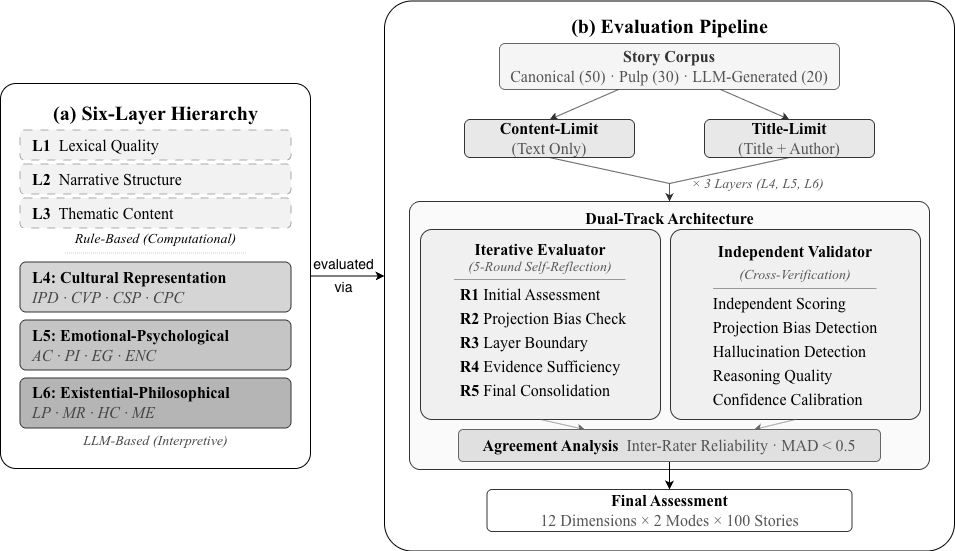}
\caption{SAGE framework architecture. (a)~Six-layer hierarchy: Layers 1--3 assess textual properties through rule-based metrics (dashed); Layers 4--6 evaluate interpretive dimensions through LLM-based analysis (solid). (b)~Evaluation pipeline: each story is assessed in two modes (content-limit and title-limit) across three layers via a dual-track architecture comprising a five-round iterative evaluator and an independent validator, with inter-rater agreement analysis.}
\label{fig:framework}
\end{figure*}

Figure~\ref{fig:framework} illustrates the overall architecture, and Table~\ref{tab:framework_layers} summarizes the six-layer hierarchy. The modular design permits selective evaluation using layer subsets depending on analytical objectives. Our experimental validation focuses on Layers 4, 5, and 6, which span cultural, emotional-psychological, and existential-philosophical dimensions with measurable criteria grounded in established theoretical frameworks.

\begin{table}[t]
\centering
\caption{Six-layer hierarchical evaluation framework}
\label{tab:framework_layers}
\small
\resizebox{\columnwidth}{!}{%
\begin{tabular}{clp{7cm}}
\hline
\textbf{Layer} & \textbf{Focus} & \textbf{Key Dimensions} \\
\hline
\multicolumn{3}{l}{\textit{Rule-Based Evaluation (Computational Metrics)}} \\
\hline
L1 & Lexical Quality & Vocabulary richness, syntactic complexity, rhetorical density \\
L2 & Narrative Structure & Entity coherence, emotional arc trajectories, event sequence patterns \\
L3 & Thematic Content & Topic diversity, semantic networks, conceptual structure \\
\hline
\multicolumn{3}{l}{\textit{LLM-Based Evaluation (Interpretive Analysis)}} \\
\hline
L4 & Cultural & \textbf{IPD}: Intersectional Power Dynamics; \\
 & Representation & \textbf{CVP}: Cultural Voice \& Perspective; \\
 & & \textbf{CSP}: Cultural Specificity; \\
 & & \textbf{CPC}: Cultural Pattern Complexity (\emph{habitus}) \\
\hline
L5 & Emotional- & \textbf{AC}: Affective Complexity; \\
 & Psychological & \textbf{PI}: Psychological Interiority; \\
 & & \textbf{EG}: Emotional Granularity; \\
 & & \textbf{ENC}: Emotional-Narrative Coherence \\
\hline
L6 & Existential- & \textbf{LP}: Life Philosophy (\emph{Weltanschauung}); \\
 & Philosophical & \textbf{MR}: Moral Reflection (\emph{phronesis}); \\
 & & \textbf{HC}: Human Condition (\emph{conditio humana}); \\
 & & \textbf{ME}: Meaning Exploration (\emph{hermeneutica}) \\
\hline
\end{tabular}
}
\end{table}

\subsection{Rule-Based Evaluation Layers (L1-L3)}

The lower three layers employ computational metrics to assess textual properties amenable to algorithmic measurement, including vocabulary richness, syntactic complexity, entity coherence, emotional arc trajectories, topic diversity, and semantic network structure. These layers provide efficient, reproducible assessment of surface and discourse-level features. As the present study focuses on validating the interpretive layers (L4--L6), the rule-based layers are summarized in Table~\ref{tab:framework_layers} and their experimental validation is left for future work.

\subsection{LLM-Based Evaluation Layers (L4-L6)}

The upper three layers employ large language models to assess interpretive dimensions requiring cultural knowledge, emotional intelligence, and philosophical sophistication. All three layers form the focus of our experimental validation, each grounded in established scholarly traditions.

\subsubsection{Layer 4: Cultural Representation}

Layer 4 evaluates how texts represent culture, power structures, and social relations. The layer draws on Bourdieu's field theory and cultural capital, Said's critique of Orientalism, Spivak's work on subaltern representation, Geertz's concept of thick description, and structural anthropology, providing systematic criteria for assessing cultural representation independently of emotional depth or philosophical content. Four dimensions operationalize this layer; Table~\ref{tab:l4_exemplars} positions representative works across the resulting dimension space.

\textbf{Intersectional Power Dynamics (IPD)} examines the degree to which texts engage with power as a structuring force in social relations. Grounded in Bourdieu's theory of cultural capital alongside Marxist, feminist, and postcolonial frameworks, the dimension assesses power relations across social axes including class, gender, race, kinship, occupation, age, religion, and institutional position. High scores require textual evidence of sophisticated treatment of power asymmetries, symbolic violence, or structural constraints rather than superficial acknowledgment of social categories; power may manifest through single or multiple axes depending on narrative context.

\textbf{Cultural Voice and Perspective (CVP)} evaluates the epistemic positioning of narrative voice relative to represented cultures, distinguishing insider perspectives that demonstrate cultural authority from outsider gazes marked by epistemic distance or exoticization. Drawing on Said's critique of Orientalism~\cite{said1978orientalism} and Spivak's work on subaltern representation~\cite{spivak1988subaltern}, the dimension assesses textual markers including linguistic register, deployment of untranslated terms, focalization strategies, and treatment of cultural knowledge as given versus as strange. CVP operates independently of authorial identity.

\textbf{Cultural Specificity (CSP)} quantifies the density and precision of cultural particularity across temporal, spatial, social, and material domains. Informed by Geertz's concept of thick description~\cite{geertz1973interpretation}, the dimension assesses whether representations are grounded in concrete cultural detail: precise temporal markers, geographic localization, attention to ritual and material practice, and deployment of culturally embedded objects. CSP captures representational thickness independently of whether cultural details serve emotional engagement or philosophical argumentation.

\textbf{Cultural Pattern Complexity (CPC, \emph{habitus})} evaluates whether a text engages with multiple cultural logics simultaneously or operates within a single internally coherent framework. The dimension assesses four properties: whether social relationships exhibit multi-layered structural logic or reduce to binary oppositions; whether conflicts reflect institutional forces or personalized antagonisms; whether texts demonstrate critical reflexivity toward represented cultural systems; and whether cultural patterns exhibit archetypal depth or genre-conventional simplification.

\begin{table}[t]
\centering
\caption{L4 exemplar matrix. H\,=\,High, M\,=\,Medium, L\,=\,Low.}
\label{tab:l4_exemplars}
\small
\setlength{\tabcolsep}{5pt}
\resizebox{\columnwidth}{!}{%
\begin{tabular}{p{3.6cm}lcccc}
\hline
\textbf{Work} & \textbf{Form} & \textbf{IPD} & \textbf{CVP} & \textbf{CSP} & \textbf{CPC} \\
\hline
\emph{Things Fall Apart} (Achebe) & Novel & H & H & H & H \\
\emph{Heart of Darkness} (Conrad) & Novel & H & L & M & M \\
\emph{The Great Gatsby} (Fitzgerald) & Novel & L & L & H & L \\
\emph{Animal Farm} (Orwell) & Novel & H & L & L & L \\
``Everyday Use'' (Walker) & Short & H & H & H & H \\
``Lady with the Dog'' (Chekhov) & Short & L & L & H & L \\
``The Dead'' (Joyce) & Short & M & H & H & H \\
``Tl\"{o}n, Uqbar...'' (Borges) & Short & L & L & M & H \\
\hline
\end{tabular}
}
\end{table}

Table~\ref{tab:l4_exemplars} demonstrates that all four dimensions are empirically non-redundant. IPD and CSP vary independently: Orwell's \emph{Animal Farm} maximizes power visibility while stripping cultural specificity to serve allegory, whereas Fitzgerald's \emph{The Great Gatsby} renders Jazz Age New York with ethnographic precision while treating aspiration rather than power asymmetry as its central subject. CVP and IPD also dissociate: Conrad's \emph{Heart of Darkness} and Achebe's \emph{Things Fall Apart} share comparable IPD scores while occupying opposite CVP positions, since Conrad renders Africa through a European observer's consciousness while Achebe renders the same colonial encounter from within Igbo life. CPC can reach high values with low IPD, as in Borges's ``Tl\"{o}n, Uqbar, Orbis Tertius,'' which stages the encounter of incommensurable epistemic systems without organizing them through power hierarchy. Sarlo's analysis of Borges notes that his free movement between cultural traditions is precisely what post-colonial critics have faulted as evasion of power relations~\cite{sarlo2007jorge}, a critical disagreement that itself confirms CPC and IPD as distinct dimensions. Short fiction spans the same dimensional range as long-form novels: ``Everyday Use'' achieves the full high profile within a compressed form, ``The Lady with the Dog'' demonstrates high CSP within a single cultural logic, and ``The Dead'' generates CPC from internal tension among cultural frameworks operating within a single social world.

\subsubsection{Layer 5: Emotional-Psychological Representation}

Layer 5 evaluates how texts represent emotions, psychological states, and affective experiences. The layer draws on affect theory (Sedgwick, Berlant, Ngai), narrative psychology (Cohn, Bruner, Bakhtin), Barrett's emotion granularity research, and James Wood's notion of motivated emotion. Four dimensions operationalize this layer; Table~\ref{tab:l5_exemplars} positions representative works across the resulting dimension space. A key property of this layer is that the four dimensions do not co-vary uniformly: a text may achieve high affective complexity with minimal psychological interiority, or produce emotionally coherent narrative while employing coarse-grained emotional language, as the table illustrates.

\textbf{Affective Complexity (AC)} measures the multiplicity, contradiction, and evolution of emotional states as represented in narrative. Grounded in affect theory~\cite{sedgwick1995shame,berlant2020cruel,ngai2009ugly}, the dimension evaluates whether emotions appear as simple singular affects or exhibit layered, contradictory, and evolving patterns. High AC manifests through coexisting contradictory emotions, affective ambiguity, and transformation across narrative progression. The dimension focuses on structural properties of emotional representation rather than intensity or explicitness of expression: Hemingway's formulation that ``the dignity of movement of an iceberg is due to only one-eighth of it being above water'' captures the principle that affective complexity can be fully present while remaining structurally submerged.

\textbf{Psychological Interiority (PI)} assesses the depth and accessibility of characters' inner psychological worlds, evaluating the degree to which narrative grants access to thoughts, motivations, self-awareness, and mental processes beyond observable behavior. Drawing on narrative psychology~\cite{cohn1978transparent}, the dimension distinguishes psychological depth from behavioral description. High PI manifests through interior monologue, free indirect discourse, psychological focalization, and related techniques revealing characters' reasoning, memory, fantasy, and conflicting desires.

\textbf{Emotional Granularity (EG)} evaluates the precision and differentiation of emotional vocabulary and concepts. Grounded in Barrett's emotion granularity research~\cite{barrett2017emotions}, the dimension assesses whether texts partition the emotional space into fine-grained distinctions (grief versus melancholy, wistfulness versus nostalgia, resentment versus disappointment) or apply broad undifferentiated categories. Granularity can operate at the lexical level through precise emotion terms or at the structural level through behavioral and situational markers that specify emotional states without naming them explicitly.

\textbf{Emotional-Narrative Coherence (ENC)} evaluates whether emotional moments arise organically from narrative situation and character or appear unmotivated and manipulative. Grounded in New Critical concepts of organic unity and James Wood's notion of motivated emotion~\cite{wood2008fiction}, the dimension assesses logical-causal relationships between emotions and their narrative contexts. High coherence requires clear antecedent motivation in plot or character development and proportionality to narrative stakes; the dimension identifies sentimentality, melodrama, and unearned pathos as markers of low coherence.

\begin{table}[t]
\centering
\caption{L5 exemplar matrix. H\,=\,High, M\,=\,Medium, L\,=\,Low. $^\dagger$EG operates at narrative apparatus level, not character self-report.}
\label{tab:l5_exemplars}
\small
\setlength{\tabcolsep}{5pt}
\resizebox{\columnwidth}{!}{%
\begin{tabular}{p{3.8cm}lcccc}
\hline
\textbf{Work} & \textbf{Form} & \textbf{AC} & \textbf{PI} & \textbf{EG} & \textbf{ENC} \\
\hline
``Hills Like White Elephants'' (Hemingway) & Short & H & L & L & H \\
``The Metamorphosis'' (Kafka) & Short & M & L & L & H \\
``Runaway'' (Munro) & Short & H & H & H & H \\
``Miss Brill'' (Mansfield) & Short & M & H & H$^\dagger$ & H \\
``Cathedral'' (Carver) & Short & L$\to$H & L & L & H \\
\emph{Mrs. Dalloway} (Woolf) & Novel & H & H & H & H \\
\emph{In Search of Lost Time} (Proust) & Novel & H & H & M & H \\
\emph{Neapolitan Novels} (Ferrante) & Novel & H & H & H & H \\
\hline
\end{tabular}
}
\end{table}

Table~\ref{tab:l5_exemplars} reveals two key orthogonalities. First, AC and PI dissociate along the rows. Hemingway's ``Hills Like White Elephants'' stages extreme emotional complexity (love, coercion, resentment, and grief simultaneously active between two characters who cannot speak directly) while providing no access to either character's inner states: AC is high, PI is near zero. Woolf's \emph{Mrs. Dalloway} achieves high scores on both dimensions simultaneously through stream-of-consciousness technique. Kafka's \emph{The Metamorphosis} occupies a third configuration: Gregor's inner life is partially accessible but estranged, his emotions rendered as functional states (``he felt guilty,'' ``was troubled'') without psychological depth, yielding medium AC and low PI. Second, EG and ENC dissociate along the Cathedral column: Carver's narrator employs deliberately flat emotional vocabulary throughout (``fine,'' ``I guess,'' ``I don't know'') yet achieves high ENC because the narrator's transformation is fully motivated by the marriage, the defensiveness, and Robert's patience that the narrative has established. The ``Miss Brill'' entry registers an analytically distinct case in which EG operates as a dramatic mechanism: the story's precision lies in the narrative apparatus, which renders the exact emotional state that the character systematically misnames, making EG failure itself the dramatic subject. Wood has observed of Ferrante that her fiction names emotional states that readers recognize from experience but have rarely seen articulated in prose~\cite{wood2013women}, a formulation that captures the operational definition of high EG with greater precision than any theoretical statement.

\subsubsection{Layer 6: Existential-Philosophical Representation}

Layer 6 evaluates philosophical depth and engagement with existential themes addressing fundamental questions of human existence. The layer draws on existentialist philosophy (Heidegger, Sartre, Camus), moral philosophy (Levinas, MacIntyre), philosophical anthropology (Arendt, Nussbaum), and hermeneutic traditions (Ricoeur, Gadamer). Four dimensions operationalize this layer; Table~\ref{tab:l6_exemplars} positions representative works across the resulting dimension space.

\textbf{Life Philosophy (LP, \emph{Weltanschauung})} examines engagement with fundamental questions of human existence, authenticity, freedom, and life's meaning, assessing whether texts sustain a coherent position on how to live through narrative situation and character confrontation with existential questions. Drawing on existentialist philosophy (Heidegger's Being-towards-death, Sartre's radical freedom, Camus's absurdism) alongside Eastern philosophical traditions including Confucianism, Daoism, and Buddhism~\cite{heidegger2010being,sartre2022being,camus1969myth}, the dimension distinguishes genuine philosophical exploration embodied through lived experience from didactic insertion of philosophical statements disconnected from narrative.

\textbf{Moral Reflection (MR, \emph{phronesis})} assesses depth of ethical inquiry, moral complexity, and engagement with questions of right conduct and responsibility. Grounded in moral philosophy (Levinas's ethics of alterity, MacIntyre's virtue ethics, Confucian ethics of reciprocity)~\cite{levinas1979totality,macintyre2013after}, the dimension evaluates whether texts present moral dilemmas with genuine complexity or reduce ethical questions to clear binaries. High MR manifests through situations where competing moral frameworks yield conflicting obligations, characters confront irreducible ambiguity without resolution, or narrative structure questions conventional moral certainties. The dimension captures sophistication of moral thinking rather than moral content: morally transgressive texts may demonstrate high MR by the seriousness with which they question ethical assumptions.

\textbf{Human Condition (HC, \emph{conditio humana})} evaluates exploration of universal aspects of human experience including mortality, suffering, finitude, vulnerability, and the search for meaning in the face of existential limits. Based on philosophical anthropology (Arendt's analysis of human plurality and natality, Nussbaum's capabilities approach)~\cite{arendt2019human,nussbaum2003upheavals}, the dimension assesses whether texts engage meaningfully with constraints and possibilities that define human existence, treating them as fundamental conditions rather than plot devices. The criterion is not whether existential limits appear in a text but whether they are thematized as conditions shaping human experience and meaning-making.

\textbf{Meaning Exploration (ME, \emph{hermeneutica})} examines how narratives construct, question, or subvert frameworks of meaning and interpretive horizons. Drawing on hermeneutic philosophy (Ricoeur's narrative identity, Gadamer's fusion of horizons)~\cite{ricoeur1980narrative,gadamer2013truth}, the dimension assesses whether texts thematize the construction of meaning itself: how individuals create coherent narratives from experience, how interpretive frameworks shape understanding, and how meaning proves fragile or contested. High ME does not require that meaning be affirmed; the dimension is equally high when meaning is systematically undermined, provided the construction and contingency of meaning are the narrative's philosophical subject.

\begin{table}[t]
\centering
\caption{L6 exemplar matrix. H\,=\,High, M\,=\,Medium, L\,=\,Low. $^*$LP structurally inaccessible (clones possess no future). $^\dagger$ME collapses at death but constitutes the story's subject throughout. $^\ddagger$MR suspended: narrative refuses moral adjudication.}
\label{tab:l6_exemplars}
\small
\setlength{\tabcolsep}{4pt}
\resizebox{\columnwidth}{!}{%
\begin{tabular}{p{3.8cm}lcccc}
\hline
\textbf{Work} & \textbf{Form} & \textbf{LP} & \textbf{MR} & \textbf{HC} & \textbf{ME} \\
\hline
\emph{The Stranger} (Camus) & Novel & H & L & H & H \\
\emph{Crime and Punishment} (Dostoevsky) & Novel & H & H & H & H \\
\emph{Germinal} (Zola) & Novel & L & M & H & L \\
\emph{Middlemarch} (Eliot) & Novel & L & H & H & M \\
\emph{Never Let Me Go} (Ishiguro) & Novel & L$^*$ & M & H & H \\
``The Guest'' (Camus) & Short & H & Susp.$^\ddagger$ & H & H \\
``A Clean, Well-Lighted Place'' (Hemingway) & Short & H & L & H & H \\
``A Hunger Artist'' (Kafka) & Short & H & L & H & H$^\dagger$ \\
``The Dead'' (Joyce) & Short & L & L-M & H & H \\
\hline
\end{tabular}
}
\end{table}

Table~\ref{tab:l6_exemplars} reveals three orthogonalities central to the layer's design. First, LP and MR dissociate in both directions. Camus's \emph{The Stranger} achieves high LP through Meursault's coherent absurdist life-position (all outcomes are equivalent, the universe is indifferent) while sustaining near-zero MR, since the novel's power depends on refusing moral evaluation throughout. Dostoevsky's \emph{Crime and Punishment} shares high LP while MR constitutes its entire subject: Raskolnikov's Napoleonic philosophy generates the moral crisis the novel then systematically prosecutes. Eliot's \emph{Middlemarch} inverts the pattern, with Dorothea Brooke demonstrating profound moral engagement through intuition rather than systematic philosophy, yielding low LP alongside high MR. Second, HC and ME dissociate: Zola's naturalism and Hemingway's ``A Clean, Well-Lighted Place'' share maximal HC, since in both, human beings face forces they cannot ultimately escape, but they diverge sharply on ME because Zola's deterministic framework forecloses meaning from within while Hemingway's old waiter constructs meaning through the maintenance of order and light against the nada. Third, LP can be structurally inaccessible rather than simply absent: in \emph{Never Let Me Go}, the clones possess no future and therefore cannot develop a life philosophy in the usual sense, yet HC is extreme and ME is high through Kathy's narration as sustained meaning-making through memory. This combination is difficult to identify through standard critical vocabulary but is immediately legible once the dimensions are named, illustrating the analytical payoff of the framework's decomposition. The short fiction entries cover the full LP and MR space, with Camus's ``The Guest'' providing a distinct MR configuration (suspended rather than low) in which Daru's principled refusal to escort a prisoner closes before the story adjudicates his choice morally.

\subsection{Dual-Mode Evaluation Framework}

To probe different aspects of LLM evaluation capacity, we implement two evaluation modes that vary the information provided to the model.

\textbf{Content-Limit Mode} provides only the story text without title, author, or metadata. This mode requires the LLM to assess literary qualities based solely on observable textual evidence without relying on memorized information about specific works or authors. Content-limit evaluation tests whether LLMs can apply generalizable critical frameworks to unfamiliar texts, analyzing cultural representation, emotional complexity, and other interpretive dimensions from textual patterns alone. Prompts in this mode explicitly instruct the model to avoid using prior knowledge and cite specific passages to ground all assessments.

\textbf{Title-Limit Mode} provides only the work's title and author without story text. This mode requires the LLM to synthesize scholarly consensus and critical reception from its training corpus. Title-limit evaluation tests whether LLMs can accurately recall and integrate expert literary criticism to produce meta-critical assessments. Prompts in this mode instruct the model to draw on academic analyses, cite specific critics or scholarly perspectives where relevant, and distinguish reliable critical consensus from speculative interpretations.

Comparing content-limit and title-limit modes reveals whether LLM evaluations derive from textual analysis, memorized critical opinions, or a combination of both sources. Systematic differences between modes indicate reliance on prior knowledge, while high agreement suggests that LLMs can perform both textual analysis and critical synthesis yielding consistent assessments.

\subsection{Dual-Track Evaluation Architecture}

To ensure reliable and interpretable assessments, we employ two evaluators with complementary roles: an iterative evaluator that performs multi-round self-reflection and an independent validator that provides cross-verification.

\subsubsection{Iterative Evaluator}

The iterative evaluator performs assessment across five structured rounds, progressively refining judgments while tracking confidence and reasoning quality. Each round addresses specific sources of bias or error through targeted self-reflection prompts.

\textbf{Round 1 (Initial Assessment):} The model receives the story (content-limit) or title/author (title-limit) along with theoretical definitions for each dimension. The model extracts relevant content, provides initial scores for all dimensions, assigns confidence ratings, and justifies assessments with specific textual evidence or critical citations.

\textbf{Round 2 (Projection Bias Check):} The model reviews Round 1 output and examines whether it imposed inappropriate assumptions about emotional expression norms, cultural contexts, or aesthetic conventions. The model revises scores if projection bias is detected or confirms Round 1 assessments if no bias is found.

\textbf{Round 3 (Layer Boundary Compliance):} The model verifies that it evaluated only the target layer's dimensions without conflating distinct layers. For Layer 4, this involves checking that cultural assessments did not incorporate emotional or philosophical considerations. For Layer 5, this involves verifying that emotional assessments did not evaluate cultural power structures or existential themes.

\textbf{Round 4 (Evidence Sufficiency):} The model examines the textual evidence or critical citations grounding each dimension score. The model adjusts confidence ratings (not scores) if evidence proves ambiguous or sparse, explicitly noting where inferences extend beyond clear textual support.

\textbf{Round 5 (Final Consolidation):} The model provides final scores with consolidated reasoning, summarizing key evidence, noting which dimensions exhibited convergence (minimal score change from Round 4 to Round 5 indicates stable assessment), and assigning overall evaluation confidence.

All rounds produce structured JSON outputs specifying dimension scores, confidence ratings, reasoning, and round-specific reflection data. This multi-round architecture addresses known LLM evaluation inconsistency by providing explicit opportunities to correct biases and verify evidence grounding.

\subsubsection{Independent Validator}

The independent validator evaluates the same texts without access to iterative evaluator outputs, providing cross-verification and hallucination detection. The validator performs single-pass assessment producing:

\begin{enumerate}
\item \textbf{Independent Scores:} Dimension-level scores and confidence ratings based solely on the validator's analysis.

\item \textbf{Validation Review:} Critical examination of whether the iterative evaluator's reasoning exhibits projection bias, hallucination, layer boundary violations, or unsupported inferences. The review identifies specific issues with severity ratings.

\item \textbf{Agreement Analysis:} Comparison between validator scores and iterative evaluator final scores, noting dimensions with strong agreement vs. significant discrepancies.

\item \textbf{Trust Level:} Overall qualitative assessment indicating confidence in the iterative evaluation reliability.
\end{enumerate}

Measuring agreement between iterative evaluator and independent validator provides inter-rater reliability evidence. High agreement suggests that both evaluators converge on similar interpretations when applying the same theoretical frameworks. Systematic disagreements on specific dimensions or text types reveal where LLM evaluation proves less reliable.

\section{Experiments}

\subsection{Dataset}

We validate the framework through systematic evaluation of 100 short stories across three quality categories designed to test discriminative capacity. The stratified corpus enables empirical testing of whether LLM-based evaluation can distinguish meaningful quality differences while controlling for potential confounds including genre, period, and provenance. To control for length effects, all stories fall within the 2,000--8,000 word range characteristic of the short fiction form.

\subsubsection{Canonical Literature Corpus}

The canonical subset comprises 50 short stories representing works consistently recognized by academic institutions, literary anthologies, and critical scholarship. Authors include Chekhov, Kafka, Joyce, Faulkner, Borges, Lu Xun, García Márquez, Hemingway, Poe, Woolf, and others whose works appear regularly in university syllabi and literary criticism. Publication dates span 1835 to 1990, encompassing major literary movements including Romanticism, Realism, Modernism, and Postmodernism. Geographic diversity includes Russian, European, North American, Latin American, and Japanese literary traditions. This heterogeneity ensures the framework encounters varied narrative techniques, thematic concerns, and cultural contexts characteristic of the literary canon.

\subsubsection{Pulp Fiction Corpus}

The pulp fiction subset comprises 30 short stories published between 1880 and 1950 in commercial genre magazines. Works drawn from publications including \textit{All-Story}, \textit{Weird Tales}, and \textit{Black Mask} represent adventure pulp, horror pulp, western pulp, detective fiction, and early science fiction. Authors include Burroughs, Lovecraft, Grey, and others writing for mass-market periodicals. This corpus provides a controlled comparison category distinguished from canonical literature by critical reception, institutional positioning, and commercial publication context. Pulp fiction exhibits technical narrative competence and reader engagement while typically employing genre conventions, formulaic plot structures, and commercial narrative strategies.

\subsubsection{LLM-Generated Story Corpus}

The LLM-generated subset comprises 20 short stories selected from the lars76/story-evaluation-llm dataset on Hugging Face~\cite{huggingface-story-eval}, which contains narratives generated by various open-source LLM models including Mistral variants, Gemma variants, and Llama variants. The corpus exhibits intentional quality stratification based on human evaluation scores from the source dataset: 10 stories received high-quality ratings (overall scores 3.58--3.78 on 5-point scale), 5 stories exhibit average quality (overall score 3.40), and 5 stories demonstrate low quality (overall scores 2.56--2.58). This stratified distribution enables testing whether LLM-based evaluation can discriminate quality levels within AI-generated content and identify distinctive characteristics of synthetic narratives across different generative models.

\begin{table}[t]
\centering
\caption{Evaluation corpus composition and distribution}
\label{tab:corpus_composition}
\small
\resizebox{\columnwidth}{!}{%
\begin{tabular}{lccl}
\toprule
\textbf{Category} & \textbf{Stories} & \textbf{Evaluations} & \textbf{Quality Characteristics} \\
\midrule
Canonical & 50 & 300 & Academically recognized, \\
          &    &     & regularly anthologized \\
\midrule
Pulp Fiction & 30 & 180 & Commercial periodicals, \\
             &    &     & genre conventions \\
\midrule
LLM-Generated & 20 & 120 & Multi-model generated, \\
              &    &     & stratified quality \\
\midrule
\textbf{Total} & \textbf{100} & \textbf{600} & \textbf{3 layers $\times$ 2 modes} \\
\bottomrule
\end{tabular}
}
\vspace{0.1cm}

\small
\textit{Notes:} Each story evaluated across 3 analytical layers (Cultural, Emotional, Existential) in 2 modes (content-limit, title-limit). Total: 100 stories $\times$ 3 layers $\times$ 2 modes = 600 evaluations. LLM-generated stories include high (n=10), average (n=5), and low (n=5) quality tiers.
\end{table}

The three-category design provides complementary quality contrasts. Comparing canonical literature to pulp fiction tests whether the framework discriminates works distinguished by critical reception and institutional canonization. Comparing human-authored stories (canonical and pulp) to LLM-generated narratives tests whether the framework identifies distinctive characteristics of AI-generated content. Comparing quality tiers within LLM-generated stories tests whether the framework tracks deliberate quality variation. Together these comparisons provide converging evidence for discriminative validity.

\subsection{Evaluation Procedure}

The experimental validation encompasses three analytical layers: Layer 4 (Cultural Representation), Layer 5 (Emotional-Psychological Representation), and Layer 6 (Existential-Philosophical Representation). As described in Methodology, each story undergoes evaluation in two modes (content-limit and title-limit) by two evaluators (iterative evaluator with 5-round reflection and independent validator). Table~\ref{tab:evaluation_matrix} summarizes the complete evaluation matrix.

\begin{table}[t]
\centering
\caption{Complete evaluation matrix across three SAGE layers}
\label{tab:evaluation_matrix}
\small
\resizebox{\columnwidth}{!}{%
\begin{tabular}{lcccc}
\toprule
\textbf{Layer} & \textbf{Dimensions} & \textbf{Stories} & \textbf{Modes} & \textbf{Evaluations} \\
\midrule
Layer 4 (Cultural) & 4 & 100 & 2 & 200 \\
Layer 5 (Emotional) & 4 & 100 & 2 & 200 \\
Layer 6 (Existential) & 4 & 100 & 2 & 200 \\
\midrule
\textbf{Total} & \textbf{12} & \textbf{100} & \textbf{2} & \textbf{600} \\
\bottomrule
\end{tabular}
}
\vspace{0.1cm}

\small
\textit{Notes:} Each evaluation includes iterative assessment (5 rounds) + independent validation (1 round). All three layers evaluate identical story set. Success rate: 600/600 (100\%). Dimensions: L4 (Intersectional Power Dynamics, Cultural Voice \& Perspective, Cultural Specificity, Cultural Pattern Complexity); L5 (Affective Complexity, Psychological Interiority, Emotional Granularity, Emotional-Narrative Coherence); L6 (Life Philosophy, Moral Reflection, Human Condition, Meaning Exploration).
\end{table}

\subsection{Implementation Details}

All LLM-based evaluations employ GPT-5-mini (OpenAI) with reasoning effort parameter set to ``medium'' to balance analytical depth and response consistency. The model supports maximum output tokens of 128,000, sufficient for processing complete short stories with detailed evaluation prompts. Each evaluation produces structured JSON outputs parsed to extract dimension scores, confidence ratings, textual evidence citations, and reasoning summaries.

\subsubsection{Prompt Engineering}

Prompts incorporate explicit theoretical frameworks as detailed in Methodology. For Layer 4 (Cultural Representation), prompts reference Bourdieu's field theory, Said's Orientalism, Geertz's thick description, and structural anthropology. For Layer 5 (Emotional-Psychological Representation), prompts reference affect theory (Sedgwick, Berlant, Ngai), narrative psychology (Cohn, Bruner, Bakhtin), emotion granularity research (Barrett), and New Critical concepts of organic unity. For Layer 6 (Existential-Philosophical Representation), prompts reference existentialist philosophy, moral philosophy, philosophical anthropology, and hermeneutic frameworks as specified in the theoretical foundation.

Prompts establish core evaluation principles including evidence sovereignty (all scores must cite textual evidence), strict layer boundaries (evaluate only target layer dimensions), and theoretical consistency (apply frameworks uniformly across texts). Prompts explicitly instruct models to avoid common pitfalls: conflating explicitness with depth, imposing evaluator's emotional or cultural norms on texts, crossing layer boundaries, and making inferences unsupported by evidence. The complete prompt templates, evaluation scripts, dataset, and results are available at \url{https://github.com/tisage/sage-eval}.

\subsubsection{Example Prompts}

\begin{promptbox}[title={\small Layer 4, Content-Limit Mode, Round 1}]
\small
You are evaluating the cultural representation quality of a literary work based solely on textual evidence. You have been provided with the story text but no information about the title, author, or publication context.

Your task is to assess four dimensions of cultural representation:

\textbf{Intersectional Power Dynamics (IPD):} Grounded in Bourdieu's theory of cultural capital alongside Marxist, feminist, and postcolonial frameworks, evaluate complexity and configuration of power relations as they operate through social structures. Power may be organized along various social axes including class, gender, race/ethnicity, kinship, occupation, age, religion, institutional position. Focus on: how power is distributed, negotiated, and legitimated through social structures; interaction between forms of capital (economic, cultural, symbolic); structural mechanisms (institutions, norms, hierarchies); complexity from simple binary oppositions to multi-layered structural tensions.

[Additional dimensions: CVP, CSP, CPC with similar theoretical grounding...]

\textbf{CRITICAL:} Base your evaluation ONLY on what is present in the text itself. Do not use any prior knowledge about literary works, authors, or cultural contexts beyond what the text explicitly provides. Cite specific passages to support all claims.

Provide scores (1.0--5.0) and confidence ratings (1--5) for each dimension with detailed reasoning.
\end{promptbox}


\begin{promptbox}[title={\small Layer 4, Title-Limit Mode, Round 1}]
\small
You are conducting a meta-critical analysis of a literary work's cultural representation quality based on scholarly consensus and critical reception.

You have been provided with the work's title and author but NOT the text itself. Your task is to synthesize assessments from literary critics, scholars, and academic analyses of this specific work.

For each of the four cultural dimensions (IPD, CVP, CSP, CPC), draw on your knowledge of how critics and scholars have evaluated this work. Consider:

\textbf{Critical Consensus:} What do major literary critics and scholars say about this work's treatment of power dynamics, cultural positioning, cultural specificity, and structural sophistication? Cite specific critics or critical schools where relevant.

\textbf{Literary Historical Context:} How has this work's cultural representation been assessed across different critical periods? Has critical evaluation evolved over time?

\textbf{Scholarly Reliability:} Distinguish between well-grounded critical consensus supported by extensive analysis and more speculative or partial interpretations.

\textbf{CRITICAL:} Base your assessment on the weight of scholarly consensus. If critical opinion is divided, note the division and explain the grounds of disagreement. Do not impose personal judgments where scholarly opinion provides clear guidance.

Provide dimension scores (1.0--5.0) based on critical consensus with detailed reasoning explaining which scholarly perspectives informed your assessment.
\end{promptbox}

\begin{table}[t]
\centering
\caption{Implementation specifications}
\label{tab:implementation_specs}
\small
\resizebox{\columnwidth}{!}{%
\begin{tabular}{lc}
\hline
\textbf{Parameter} & \textbf{Value} \\
\hline
\multicolumn{2}{l}{\textit{LLM Configuration}} \\
Model & GPT-5-mini \\
Provider & OpenAI \\
Reasoning effort & medium \\
Max output tokens & 128,000 \\
Output format & Structured JSON \\
\hline
\multicolumn{2}{l}{\textit{Scoring Configuration}} \\
Dimension score range & 1.0--5.0 \\
Confidence rating range & 1--5 \\
Convergence threshold & $< 0.3$ (Round 4$\rightarrow$5) \\
Inter-rater agreement threshold & MAD $< 0.5$ \\
\hline
\multicolumn{2}{l}{\textit{Evaluation Structure}} \\
Iterative evaluator rounds & 5 \\
Independent validator rounds & 1 \\
\hline
\end{tabular}%
}
\end{table}

\subsection{Statistical Analysis}

We analyze evaluation reliability, discriminative validity, and dimensional independence through multiple complementary analyses organized around the four research questions.

For evaluation reliability (RQ1), we track score trajectories across the five iterative rounds for each dimension and story. Dimensions exhibiting minimal score change (absolute difference $< 0.3$) from Round 4 to Round 5 are classified as convergent. We calculate convergence rates by dimension, layer, mode, and story category. Inter-rater reliability between iterative evaluator and independent validator is quantified through mean absolute difference (MAD), with agreement rates meeting the MAD $< 0.5$ threshold indicating acceptable reliability.

For discriminative validity (RQ2), we employ independent samples t-tests to compare mean scores across the three story categories (canonical vs.\ pulp vs.\ LLM-generated) for each layer. Effect sizes are reported through Cohen's $d$ with standard interpretation thresholds: small ($d$=0.2), medium ($d$=0.5), large ($d$=0.8). One-way ANOVA tests assess overall genre effects, followed by post-hoc pairwise comparisons with Bonferroni correction. Dimension-level analysis identifies which specific dimensions contribute most to category discrimination.

For evaluation robustness (RQ3), we compare scores between content-limit and title-limit modes using paired samples t-tests. Mode effects reveal whether LLM evaluations rely primarily on textual analysis, memorized critical consensus, or both sources.

For dimensional independence (RQ4), we compute Pearson and Spearman correlation coefficients between all pairs of Layer 4, Layer 5, and Layer 6 scores. Moderate correlations ($r < 0.7$) support dimensional independence, while high correlations would suggest substantial overlap requiring framework revision. We also compare cross-layer score distributions to test whether the three analytical dimensions exhibit differential genre sensitivity.

The evaluation framework produces both quantitative scores enabling statistical analysis and qualitative rationales permitting examination of how LLM assessment responds to specific literary features. This dual output supports both hypothesis testing and interpretive analysis of evaluation patterns.
\section{Results}
\label{sec:results}

\subsection{System Performance and Reliability}

The SAGE framework demonstrates exceptional system-level performance across 600 evaluations. Table~\ref{tab:system_reliability} summarizes key reliability metrics.

\begin{table}[t]
\centering
\caption{System reliability and quality assurance metrics}
\label{tab:system_reliability}
\small
\resizebox{\columnwidth}{!}{%
\begin{tabular}{lcc}
\toprule
\textbf{Metric} & \textbf{Value} & \textbf{Interpretation} \\
\midrule
\multicolumn{3}{l}{\textit{Evaluation Success}} \\
Total evaluations & 600 & Perfect dataset balance \\
Success rate & 100.0\% & All evaluations completed \\
Parsing success & 100.0\% & All outputs valid JSON \\
\midrule
\multicolumn{3}{l}{\textit{Convergence \& Consistency}} \\
Convergence rate & 98.8\% & Score stability (R4$\rightarrow$R5) \\
Peer agreement & $>$94\% & Inter-rater reliability \\
Mean IRR difference & $<$0.3 & Excellent agreement \\
\midrule
\multicolumn{3}{l}{\textit{Mode Robustness}} \\
Mode invariance & Max $|\Delta|$=0.05 & Content $\approx$ Title \\
\bottomrule
\end{tabular}
}
\vspace{0.1cm}

\small
\textit{Notes:} Convergence = absolute score change $<$0.3 from Round 4 to Round 5. Peer agreement = primary-validator difference $<$0.5. IRR = inter-rater reliability. Mode invariance = maximum absolute difference between content-limit and title-limit across all layers.
\end{table}

\begin{figure*}[t]
\centering
\includegraphics[width=\textwidth]{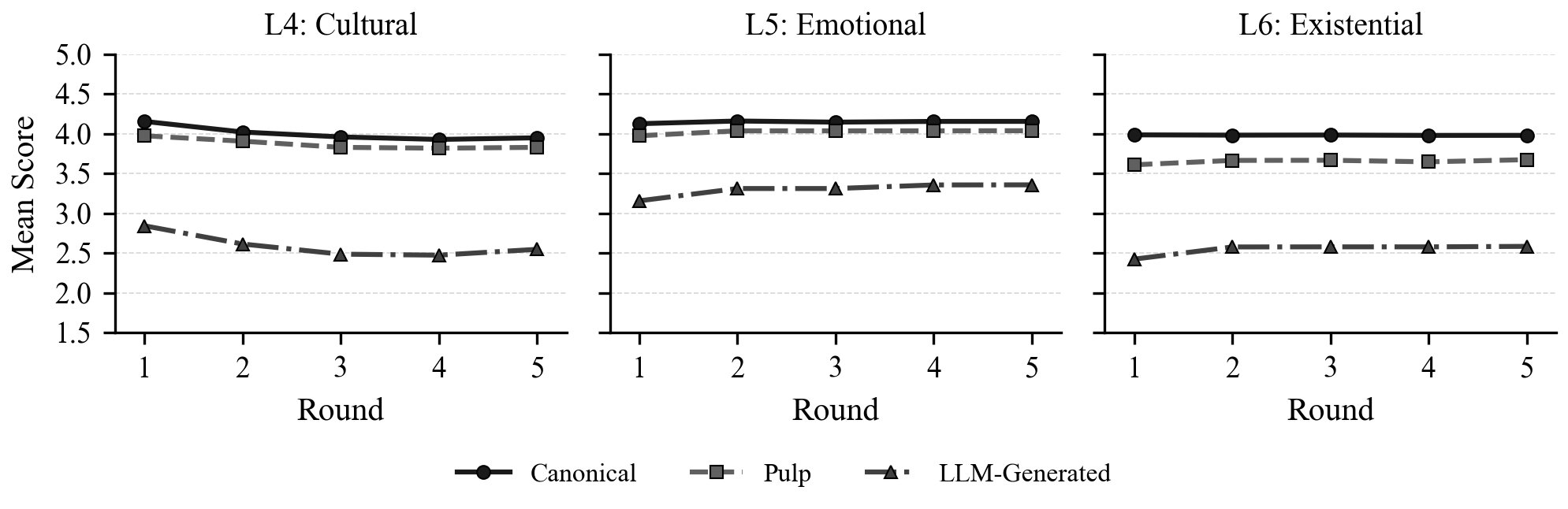}
\caption{Score convergence trajectories across five iterative rounds for each layer (L4--L6) and genre category. Scores stabilize rapidly by Round 3--4, with canonical and pulp fiction consistently scoring higher than LLM-generated stories across all layers and rounds.}
\label{fig:convergence}
\end{figure*}

All 600 evaluations completed successfully with valid JSON outputs, achieving a 100\% success rate. The five-round iterative reflection mechanism achieves 98.8\% convergence rate, with dimension scores stabilizing between Round 4 and Round 5. Figure~\ref{fig:convergence} illustrates the score convergence trajectories across all five rounds for each layer and genre category, showing rapid stabilization by Rounds 3--4. Independent validation through peer review yields greater than 94\% agreement rate across all dimensions and layers, with mean inter-rater difference below 0.3 points.

Table~\ref{tab:mode_invariance} presents mode comparison across all three layers, with maximum absolute difference of 0.05 between content-limit and title-limit modes. Paired samples t-tests confirm no significant mode effects (all $p>0.05$).

\begin{table}[t]
\centering
\caption{Mode invariance: Content-limit vs Title-limit comparison}
\label{tab:mode_invariance}
\small
\resizebox{\columnwidth}{!}{%
\begin{tabular}{lccc}
\toprule
\textbf{Layer} & \textbf{Content-Limit} & \textbf{Title-Limit} & \textbf{Difference} \\
\midrule
Layer 4 (Cultural)     & 3.63 & 3.65 & +0.02 \\
Layer 5 (Emotional)    & 3.98 & 3.93 & $-$0.05 \\
Layer 6 (Existential)  & 3.60 & 3.59 & $-$0.01 \\
\midrule
\textbf{Overall}       & \textbf{3.74} & \textbf{3.72} & \textbf{$-$0.01} \\
\bottomrule
\end{tabular}
}
\vspace{0.1cm}

\small
\textit{Notes:} Mean scores across 100 stories per mode. All differences $<$0.05 indicate mode-invariant performance. Content-limit: evaluator receives only story text (no metadata). Title-limit: evaluator receives only title + author (no text).
\end{table}

\subsection{Genre Discrimination}

\begin{figure}[t]
\centering
\includegraphics[width=\columnwidth]{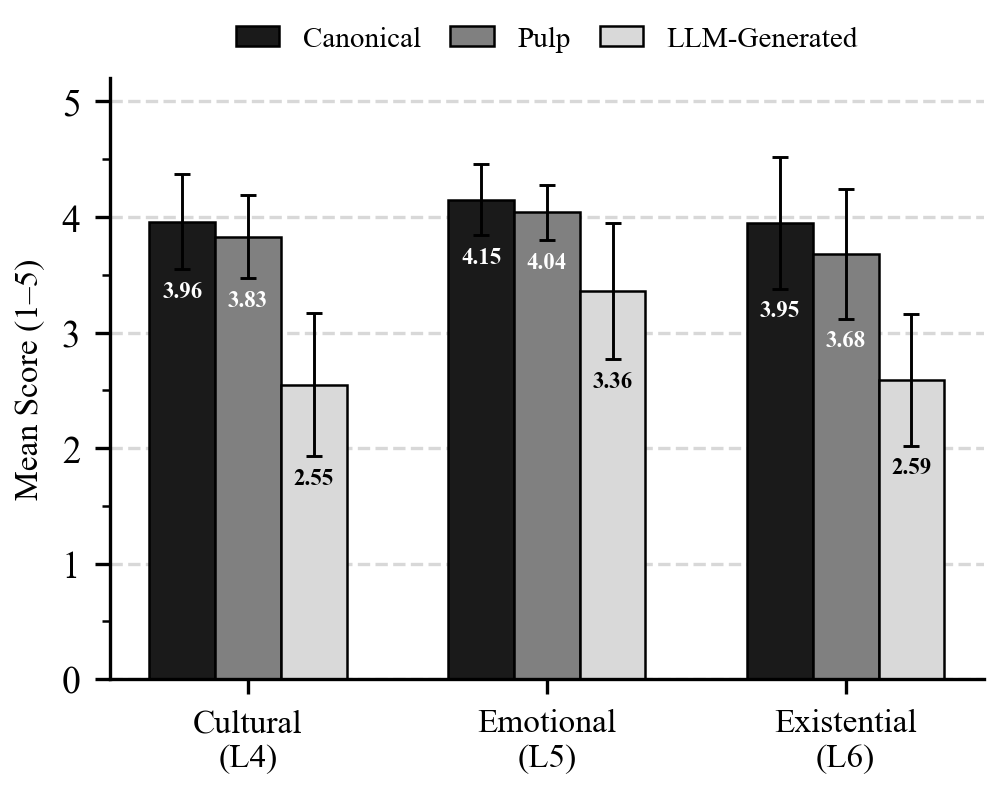}
\caption{Genre comparison across three analytical layers. Error bars indicate standard deviations. All pairwise differences between canonical and LLM-generated stories are significant at $p<0.001$.}
\label{fig:genre_comparison}
\end{figure}

Figure~\ref{fig:genre_comparison} presents genre comparison across all three analytical layers, revealing a consistent and statistically significant quality hierarchy. Detailed numerical results with standard deviations and pairwise differences are provided in Table~\ref{tab:genre_comparison_appendix} (Appendix).

\begin{figure}[t]
\centering
\includegraphics[width=\columnwidth]{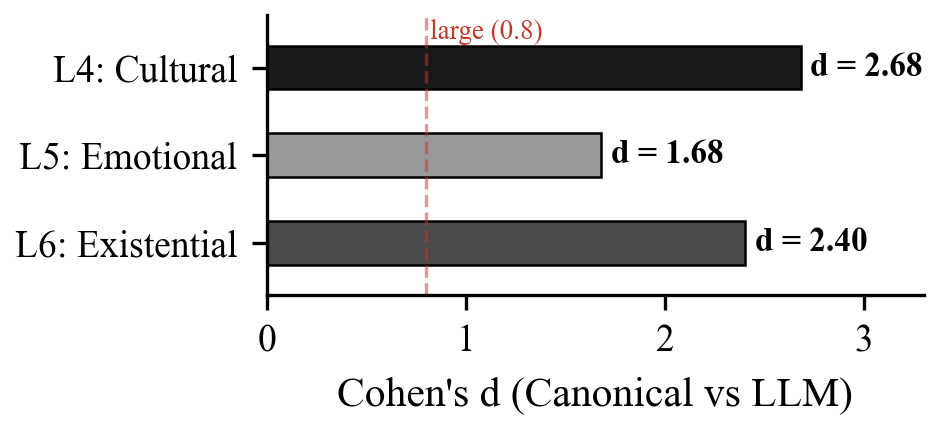}
\caption{Effect size (Cohen's $d$) for Canonical vs.\ LLM-generated stories across three layers. All values exceed the ``large effect'' threshold ($d=0.8$), indicated by the dashed red line.}
\label{fig:effect_sizes}
\end{figure}

Across all three layers, canonical literature significantly outperforms both pulp fiction and LLM-generated stories (all pairwise $p<0.001$), yielding a consistent genre ranking: Canonical (4.02) $>$ Pulp (3.85) $>$ LLM (2.83). The Canonical vs.\ LLM difference of +1.19 points (29.6\% relative gap) with very large effect sizes (Cohen's $d>1.5$ across all layers) confirms that the framework identifies fundamental quality differences between human-authored canonical literature and AI-generated narratives. Figure~\ref{fig:effect_sizes} visualizes these effect sizes across layers. The Canonical vs.\ Pulp difference remains modest (+0.17 points, 4.4\%).

\subsection{Layer-Specific Results}

\subsubsection{Layer 4: Cultural Representation}

Layer 4 achieves a mean score of 3.64 across 200 evaluations (100 stories $\times$ 2 modes). Canonical works (3.96 $\pm$ 0.41) significantly outperform pulp fiction (3.83 $\pm$ 0.36, $p<0.01$) and LLM-generated stories (2.55 $\pm$ 0.62, $p<0.001$), yielding the largest Canonical vs.\ LLM difference among the three layers (+1.41 points, 35.6\% relative gap, $d$=2.68). All four cultural dimensions (Intersectional Power Dynamics, Cultural Voice \& Perspective, Cultural Specificity, and Cultural Pattern Complexity) exhibit consistent genre hierarchy with uniformly large effect sizes.

\subsubsection{Layer 5: Emotional-Psychological Representation}

Layer 5 achieves the highest mean score (3.96) across all three layers. Canonical literature (4.15 $\pm$ 0.31) significantly outperforms LLM-generated stories (3.36 $\pm$ 0.59, $p<0.001$, $d$=1.68), but the 19.1\% relative gap is the smallest among the three layers. Pulp fiction (4.04 $\pm$ 0.24) approaches canonical performance with only 0.11-point difference (2.7\%), the smallest Canonical vs.\ Pulp gap across all layers.

\subsubsection{Layer 6: Existential-Philosophical Representation}

Layer 6 demonstrates strong genre discrimination comparable to Layer 4. Canonical works (3.95 $\pm$ 0.57) substantially exceed LLM-generated stories (2.59 $\pm$ 0.57) by +1.36 points (34.5\% relative gap, $p<0.001$, $d$=2.40). The Canonical vs.\ Pulp difference (+0.27, 6.8\%) is the largest among the three layers, with pulp fiction (3.68 $\pm$ 0.56) achieving moderate existential engagement.

\subsection{Cross-Layer Correlation}

To assess whether the three analytical layers capture genuinely distinct quality dimensions (RQ4), we compute pairwise correlation coefficients across all 200 story-mode observations. Table~\ref{tab:cross_layer_correlation} presents the results.

\begin{table}[t]
\centering
\caption{Cross-layer correlation analysis (Pearson $r$)}
\label{tab:cross_layer_correlation}
\small
\resizebox{\columnwidth}{!}{%
\begin{tabular}{lccc}
\toprule
\textbf{Layer Pair} & \textbf{Pearson $r$} & \textbf{Spearman $\rho$} & \textbf{$p$-value} \\
\midrule
L4 $\times$ L5 (Cultural $\leftrightarrow$ Emotional)     & 0.649 & 0.514 & $<$0.001 \\
L4 $\times$ L6 (Cultural $\leftrightarrow$ Existential)   & 0.683 & 0.598 & $<$0.001 \\
L5 $\times$ L6 (Emotional $\leftrightarrow$ Existential)  & 0.680 & 0.621 & $<$0.001 \\
\bottomrule
\end{tabular}
}
\vspace{0.1cm}

\small
\textit{Notes:} $N$=200 (100 stories $\times$ 2 modes). Spearman rank correlations provided as robustness check. All correlations significant at $p<0.001$.
\end{table}

All three layer pairs exhibit moderate positive correlations (Pearson $r$ = 0.649--0.683), statistically significant at $p<0.001$, yet falling below the $r>0.7$ threshold that would suggest dimensional redundancy. Spearman rank correlations ($\rho$ = 0.514--0.621) corroborate this pattern. The three layers exhibit distinct overall means: Layer 5 (Emotional, 3.96) $>$ Layer 4 (Cultural, 3.64) $>$ Layer 6 (Existential, 3.59), indicating that emotional depth is the most widespread quality dimension while cultural critique and philosophical engagement vary more substantially by genre and authorship type.

\section{Discussion}
\label{sec:discussion}

The experimental results establish that the SAGE framework achieves reliable evaluation performance while revealing systematic patterns in how different quality dimensions distinguish literary categories. We organize our interpretation around the four research questions: differential discrimination patterns address discriminative capacity (RQ2), dimensional independence validates the multi-layer decomposition (RQ4), mode invariance bears on evaluation robustness (RQ3), and the multi-round architecture contributes to reliability (RQ1). We then consider broader implications for LLM evaluation methodology and AI-generated literature.

\subsection{Differential Discrimination across Interpretive Layers}

The most striking finding concerns the differential discrimination magnitudes across the three analytical layers. Cultural representation and existential-philosophical representation exhibit very large effect sizes when comparing canonical literature to LLM-generated stories ($d$=2.68 and $d$=2.40 respectively, with approximately 35\% relative gaps), while emotional-psychological representation shows a substantially smaller though still large effect ($d$=1.68, 19.1\% gap). This pattern is not an artifact of scoring range or variance: the three layers assess the same 100 stories using the same evaluation architecture, differing only in the theoretical frameworks guiding assessment. The differential must therefore reflect genuine differences in how well current generative models reproduce different dimensions of literary quality.

A plausible interpretation draws on the distinction between pattern-reproducible and stance-requiring literary capacities. Emotional structures, including affective complexity, psychological interiority, and emotional granularity, are extensively represented in narrative training corpora. Characters experience recognizable emotions; narrators convey interiority through established literary techniques; emotional arcs follow identifiable patterns. An LLM trained on large quantities of fiction can learn to reproduce these surface patterns with moderate fidelity, achieving scores around 3.36 compared to the canonical mean of 4.15. Cultural power critique and existential-philosophical inquiry, by contrast, require capacities that pattern matching alone may not deliver: a critical stance toward the social structures being represented, original philosophical engagement with questions of human finitude and meaning, and the ability to position narrative voice within contested ideological terrain. The approximately doubled effect sizes for these dimensions suggest they represent qualitatively different challenges for generative models rather than merely harder versions of the same task.

The three layers also exhibit distinct overall score distributions: emotional representation achieves the highest mean (3.96) across all text categories, followed by cultural representation (3.64) and existential-philosophical representation (3.59). This ranking suggests that emotional depth represents a more universally present quality dimension across diverse narratives, while cultural critique and philosophical engagement constitute more specialized achievements that vary substantially by genre and authorship context. The relatively high emotional scores even among LLM-generated stories support the interpretation that affective patterns are among the most learnable aspects of literary craft from training data alone.

\subsection{Canonical Literature and Genre Fiction}

The relationship between canonical literature and pulp fiction provides a complementary perspective on what the framework measures. Across all three layers, canonical works outperform pulp fiction by only modest margins (overall +0.17 points, 4.4\%), with the smallest gap appearing in emotional representation (2.7\%) and the largest in existential-philosophical representation (6.8\%). This convergence suggests that effective emotional representation constitutes a shared competence of human authorship, achieved through authentic lived experience regardless of critical positioning or institutional recognition. The larger gap in philosophical depth indicates that sustained existential inquiry distinguishes canonical literature from commercial genre fiction more strongly than either emotional or cultural sophistication, a finding consistent with the literary critical tradition that prizes philosophical ambiguity and interpretive openness as hallmarks of canonical status.

This pattern also carries methodological significance. A coarser evaluation framework that collapsed cultural, emotional, and philosophical dimensions into a single quality score would yield a modest 4.4\% canonical-pulp difference that might appear negligible. The layered decomposition reveals that this small aggregate difference masks meaningful variation: canonical and pulp fiction converge on emotional competence while diverging on philosophical depth. Such differential sensitivity across dimensions validates the ontological decomposition motivating the SAGE architecture and suggests that aggregate quality scores obscure more than they reveal.

\subsection{Dimensional Independence and Framework Validity}

The cross-layer correlation analysis (Table~\ref{tab:cross_layer_correlation}) provides direct evidence for RQ4. All three layer pairs yield moderate Pearson correlations ($r$ = 0.649--0.683), confirming that cultural, emotional-psychological, and existential-philosophical representation capture empirically distinguishable quality dimensions. The correlations are positive, as expected: higher-quality works tend to excel across multiple dimensions, and texts with impoverished cultural representation are unlikely to demonstrate profound philosophical engagement. Yet the correlations remain well below the threshold ($r>0.7$) that would suggest the layers are measuring the same underlying construct under different labels.

The asymmetry in the correlation structure is itself informative. The strongest association links cultural and existential dimensions ($r$=0.683), consistent with the observation that sustained cultural critique and philosophical inquiry both require a critical stance toward the material being represented, a capacity that distinguishes canonical literature from both commercial genre fiction and LLM-generated narratives. The weakest association links cultural and emotional dimensions ($r$=0.649), supporting the finding that emotional competence is more broadly distributed across text categories and less dependent on the analytical sophistication required for cultural power critique. These patterns reinforce the methodological claim that collapsing distinct evaluative dimensions into a single quality metric sacrifices meaningful discriminative information.

\subsection{Mode Invariance and Evaluation Grounding}

The near-perfect mode invariance observed across all layers (maximum absolute difference of 0.05 between content-limit and title-limit modes) warrants careful interpretation. In content-limit mode, the evaluator works solely from textual evidence without knowledge of authorship or critical reception. In title-limit mode, the evaluator synthesizes scholarly consensus and critical reputation without access to the text itself. That these two radically different information sources yield statistically indistinguishable assessments admits at least two readings. The optimistic interpretation holds that LLM evaluation captures genuine literary properties accessible through either direct textual analysis or accumulated critical judgment, and that the convergence validates both the framework and the consistency of the evaluator's critical reasoning. A more cautious reading notes that even content-limit evaluation cannot fully exclude the influence of training corpus knowledge, since stylistic features of well-known works may activate learned associations without explicit authorship recognition.

The truth likely involves elements of both: LLM evaluators perform genuine textual analysis while also drawing on internalized critical knowledge, producing assessments that reflect the same underlying quality dimensions regardless of the information pathway. For practical purposes, this mode invariance is advantageous. It suggests the framework can be applied with equal confidence to texts with and without established critical reception, extending its utility beyond canonical works to contemporary and emerging literature where scholarly consensus has not yet formed.

\subsection{Implications for LLM-as-Judge Methodology}

These findings contribute to the growing literature on LLM-as-judge evaluation. Prior work has demonstrated that LLM evaluators can achieve substantial agreement with human preferences in conversational and instructional settings, while also exhibiting systematic biases including position bias, verbosity bias, and self-enhancement~\cite{zheng2023judging,li2024llms}. Our results extend this literature to literary evaluation, demonstrating that with appropriate architectural design, LLM evaluators can achieve inter-rater agreement exceeding 94\% and convergence rates of 98.8\% on interpretive tasks considerably more complex than typical NLG evaluation. The multi-round architecture proves particularly important: the structured progression from initial assessment through projection bias checking, layer boundary verification, evidence sufficiency review, and final consolidation provides explicit mechanisms for addressing known sources of LLM evaluation error that single-pass approaches cannot match.

Whether this level of reliability generalizes across different LLM families and evaluation domains remains an open empirical question. The present study employs a single model (GPT-5-mini) for all evaluations. Cross-model comparison across architecturally distinct LLM families would distinguish framework-level reliability from model-specific behavioral properties and strengthen the case for multi-round iterative evaluation as a general methodology rather than a model-dependent technique.

\subsection{Implications for AI-Generated Literature}

The systematic capability profile revealed by the framework carries implications beyond evaluation methodology. The finding that current LLMs achieve moderate competence at emotional pattern reproduction while demonstrating fundamental weaknesses in cultural power critique and existential-philosophical depth provides empirically grounded guidance for both AI development and literary assessment. For AI developers, these results indicate that improvements in surface fluency and emotional mimicry should not be mistaken for progress toward genuine literary capability, since the most distinctive achievements of canonical literature lie precisely in the dimensions where AI generation performs worst. For literary scholars and critics engaging with AI-generated content, the framework provides systematic criteria for distinguishing technically competent but philosophically shallow AI narratives from works exhibiting the critical depth and existential authenticity associated with the human literary tradition.

More broadly, the approximately 35\% relative gaps in cultural and existential dimensions raise the question of whether these weaknesses reflect fundamental architectural limitations of current language models or temporary gaps addressable through improved training and prompting strategies. The pattern-reproducibility interpretation developed above suggests the former: if cultural critique and philosophical depth require genuine critical stance and existential confrontation rather than statistical pattern matching, then scaling training data or refining prompts may yield diminishing returns. Longitudinal evaluation as generative models improve would provide empirical evidence to adjudicate this question, tracking whether the capability hierarchy identified here proves stable or shifts with architectural advances.

\section{Conclusion}
\label{sec:conclusion}

This paper introduced the SAGE framework, a hierarchical system for evaluating literary quality through systematic decomposition into interpretive dimensions assessed by large language models. Through comprehensive evaluation of 100 short stories across three analytical layers using dual-mode assessment with multi-round iterative reflection, we address four research questions and demonstrate that LLM-based literary evaluation can achieve strong measurement reliability while revealing systematic patterns distinguishing canonical literature, commercial genre fiction, and AI-generated narratives.

Regarding reliability (RQ1), the framework achieves robust performance with 100\% evaluation success, 98.8\% score convergence, and greater than 94\% inter-rater agreement, validating the multi-round architecture as an effective mechanism for grounding interpretive judgments. Regarding discriminative capacity (RQ2), the consistent genre hierarchy (Canonical $>$ Pulp $>$ LLM, all $p<0.001$) with layer-specific discrimination patterns (very large effect sizes for cultural and existential dimensions at $d>2.4$, smaller but still large effects for emotional dimension at $d$=1.68) reveals a capability profile distinguishing pattern-reproducible from stance-requiring literary capacities. Regarding evaluation robustness (RQ3), near-perfect mode invariance (maximum $|\Delta|$=0.05) demonstrates that content-based textual analysis and metadata-based critical synthesis converge on equivalent assessments. Regarding dimensional independence (RQ4), the framework addresses non-redundancy at two levels. At the within-layer level, the exemplar matrices in Tables~\ref{tab:l4_exemplars}--\ref{tab:l6_exemplars} demonstrate that dimensions within each layer dissociate in concrete texts: works can achieve high affective complexity with near-zero psychological interiority, or high emotional coherence alongside low emotional granularity, confirming that the twelve dimensions provide conceptually independent analytical purchase. At the cross-layer level, moderate correlations ($r$ = 0.649--0.683) confirm that cultural, emotional-psychological, and existential-philosophical representation constitute empirically distinguishable quality dimensions that statistical aggregation would obscure.

This work contributes to computational literary studies and LLM evaluation research in three respects. First, we provide a theoretically grounded framework that operationalizes abstract critical concepts, from Bourdieu's cultural capital to Heideggerian existential analysis, through structured prompting strategies. Construct validity is established through exemplar-grounded orthogonality analysis: each of the twelve dimensions is anchored by canonical literary contrasts demonstrating that dimensions dissociate in concrete texts, providing a theory-level argument for non-redundancy that complements the statistical cross-layer evidence. Second, we demonstrate that LLM-based evaluation with multi-round reflection, independent validation, and explicit theoretical grounding can achieve reliability comparable to expert literary criticism, establishing a scalable pathway for automated interpretive assessment beyond surface feature extraction. Third, we characterize systematic limitations of current generative models through controlled quality comparison, revealing that approximately 35\% performance gaps in cultural critique and philosophical depth indicate fundamental challenges in AI literary generation that surface fluency and emotional mimicry cannot bridge.

Several limitations warrant acknowledgment. The evaluation corpus comprises 100 English-language short stories, which restricts generalizability to other languages, literary forms, and cultural contexts. The study employs a single model (GPT-5-mini); cross-model comparison would determine whether the reliability properties observed here reflect framework-level design or model-specific behavior. The reported inter-rater agreement exceeding 94\% reflects consistency between two LLM-based evaluators rather than agreement with human experts, and validation against professional literary critics remains essential for establishing external validity. The framework validates only the interpretive layers (L4--L6), leaving integration with the rule-based computational metrics of L1--L3 for future work.

These limitations point toward several productive directions. Expanding the corpus to non-English literatures and diverse narrative forms would test the universality of the theoretical frameworks underlying each layer. Systematic cross-model comparison across LLM families would separate framework reliability from model-specific properties. Human expert benchmarking would provide the external validity that LLM-to-LLM agreement cannot supply. Longitudinal evaluation as generative models improve would track whether the capability profile identified here remains stable or shifts with architectural advances, providing empirical grounding for understanding where the boundaries of AI literary generation ultimately lie.

\bibliography{reference, new_reference}
\bibliographystyle{aaai}

\section*{APPENDIX}
\label{sec:appendix}

\subsection*{Detailed Genre Comparison}

\begin{table}[ht]
\centering
\caption{Genre comparison across three analytical dimensions (detailed)}
\label{tab:genre_comparison_appendix}
\small
\resizebox{\columnwidth}{!}{%
\begin{tabular}{lcccccc}
\toprule
\textbf{Layer} & \textbf{Canon.} & \textbf{Pulp} & \textbf{LLM} & \textbf{Can-Pulp} & \textbf{Can-LLM} & \textbf{$p$} \\
\midrule
Cultural (L4)     & 3.96 & 3.83 & 2.55 & +0.13 & \textbf{+1.41} & $<$0.001*** \\
                  & (0.41) & (0.36) & (0.62) & & & \\
\midrule
Emotional (L5)    & 4.15 & 4.04 & 3.36 & +0.11 & \textbf{+0.79} & $<$0.001*** \\
                  & (0.31) & (0.24) & (0.59) & & & \\
\midrule
Existential (L6)  & 3.95 & 3.68 & 2.59 & +0.27 & \textbf{+1.36} & $<$0.001*** \\
                  & (0.57) & (0.56) & (0.57) & & & \\
\midrule
\textbf{Average} & \textbf{4.02} & \textbf{3.85} & \textbf{2.83} & \textbf{+0.17} & \textbf{+1.19} & $<$\textbf{0.001***} \\
\bottomrule
\end{tabular}
}
\vspace{0.1cm}

\small
\textit{Notes:} Mean scores with standard deviations in parentheses. Can-Pulp/Can-LLM = score differences. Statistical significance: ***$p<0.001$. Effect sizes (Cohen's $d$) for Canonical vs LLM: L4 $d$=2.68, L5 $d$=1.68, L6 $d$=2.40 (all very large effects).
\end{table}

\subsection*{Iterative Evaluation Protocol}

Table~\ref{tab:eval_protocol} summarizes the five-round iterative evaluation protocol. All three layers share the same round structure, with layer-specific adaptations noted.

\begin{table}[ht]
\centering
\caption{Five-round iterative evaluation protocol}
\label{tab:eval_protocol}
\small
\resizebox{\columnwidth}{!}{%
\begin{tabular}{clp{5.5cm}}
\toprule
\textbf{Round} & \textbf{Focus} & \textbf{Task} \\
\midrule
R1 & Initial Assessment & Extract layer-relevant content; score all 4 dimensions (1.0--5.0) with confidence ratings and textual evidence \\
\midrule
R2 & Bias Check & \textbf{L4}: Hallucination check (fabricated cultural/historical details). \textbf{L5/L6}: Projection bias check (imposing Western emotional norms or existentialist frameworks) \\
\midrule
R3 & Layer Boundary & Verify no cross-layer contamination (e.g., L5 must not evaluate cultural power structures [L4] or philosophical claims [L6]) \\
\midrule
R4 & Evidence Sufficiency & Assess evidence quality; adjust \textit{confidence} (not scores) if evidence is ambiguous or sparse \\
\midrule
R5 & Final Consolidation & Confirm or adjust final scores; convergence criterion: $|\Delta\text{score}| < 0.3$ from R4 to R5 \\
\bottomrule
\end{tabular}
}
\end{table}

The independent validator receives the iterative evaluator's five-round output and performs single-pass cross-verification covering: (1)~independent scoring, (2)~projection bias detection, (3)~hallucination detection, (4)~reasoning quality assessment, and (5)~confidence calibration. Agreement between evaluator and validator is measured per dimension, with discrepancies $>$0.5 points flagged for review.

\subsection*{Dimension Scoring Criteria}

Each dimension is scored on a 1.0--5.0 scale. Table~\ref{tab:scoring_rubric} provides the scoring rubric applied uniformly across all layers.

\begin{table}[ht]
\centering
\caption{Scoring rubric for dimension evaluation}
\label{tab:scoring_rubric}
\small
\begin{tabular}{cp{6.5cm}}
\toprule
\textbf{Score} & \textbf{Criteria} \\
\midrule
4.5--5.0 & \textbf{Exceptional}: Rich, multi-layered representation with extensive textual evidence; sophisticated treatment demonstrating mastery \\
4.0--4.5 & \textbf{Strong}: Substantial representation with clear evidence; demonstrates depth and nuance \\
3.0--3.5 & \textbf{Moderate}: Some representation present but limited in scope or depth; evidence adequate but not extensive \\
2.0--2.5 & \textbf{Weak}: Minimal representation; surface-level treatment with sparse evidence \\
1.0--1.5 & \textbf{Absent/Negligible}: Dimension effectively absent or trivially treated; no meaningful evidence \\
\bottomrule
\end{tabular}
\end{table}

\subsection*{Prompt Template (Representative Example)}

The following presents a condensed version of the content-limit system prompt for Layer~5 (Emotional-Psychological), illustrating the prompt structure shared across all three layers. Layer~4 and Layer~6 follow the same template with layer-specific dimension definitions and theoretical frameworks.

\begin{quote}
\small
\ttfamily

\textbf{System Prompt (condensed):}

You are an expert literary-emotional analyst evaluating how texts represent emotions, psychological states, and affective experiences.

\textbf{Core Principles:}\\
1. Evidence Sovereignty: All scores must cite specific textual evidence.\\
2. Strict Layer Boundary: Evaluate ONLY emotional-psychological content. Do NOT evaluate cultural power structures (Layer~4) or philosophical themes (Layer~6).\\
3. Projection Awareness: Avoid imposing assumptions about ``how emotions should be expressed.'' Restrained expression $\neq$ shallow.\\
4. Dimension Independence: Evaluate each dimension independently.

\textbf{Mode: content-limit}\\
Evaluate based SOLELY on textual evidence. Do NOT use prior knowledge about literary works or authors.

\textbf{Dimensions:}\\
1. AC (Affective Complexity) -- Multiplicity, contradiction, and evolution of emotional states. [Sedgwick, Berlant, Ngai]\\
2. PI (Psychological Interiority) -- Depth and accessibility of characters' inner psychological worlds. [Cohn, Bruner, Bakhtin]\\
3. EG (Emotional Granularity) -- Precision and differentiation of emotional vocabulary. [Barrett]\\
4. ENC (Emotional-Narrative Coherence) -- Whether emotions arise organically from narrative situation. [New Criticism, James Wood]

\medskip

\textbf{Round 1 User Prompt (condensed):}

ROUND 1/5: EMOTIONAL CONTENT EXTRACTION

Step 1: Extract emotional content (explicit emotion words, psychological states, interior techniques, affective moments).\\
Step 2: Score each dimension (1.0--5.0) with confidence (1--5), reasoning, and evidence strength.

Output: JSON with extracted\_content, scores (per dimension: score, confidence, reasoning, evidence\_strength), overall\_score.

\medskip

\textbf{Round 2 User Prompt (condensed):}

ROUND 2/5: PROJECTION BIAS CHECK

Review Round~1. Ask critically:\\
-- Did I penalize restrained emotional expression as ``shallow''?\\
-- Did I apply Western emotional display norms?\\
-- Did I reward explicit language without checking substance?\\
Revise scores if bias detected; confirm if none found.
\end{quote}

Rounds 3--5 follow the protocol described in Table~\ref{tab:eval_protocol}, with layer-specific adaptations at each round.

\medskip
\noindent\textbf{Layer-Differentiated Round~2 Bias Mitigation.}
Round~2 is the primary site of layer-specific bias mitigation. The three layers apply distinct self-reflection strategies reflecting their different epistemological risks.

\textit{Layer~4 (Cultural)}: Because cultural-historical claims carry a distinct verification risk absent from affective or philosophical evaluation, Round~2 applies a \textit{hallucination check} rather than a projection bias check:

\begin{quote}
\small\ttfamily
ROUND 2/5: CRITICAL SELF-REFLECTION

1. Hallucination Check: Did you make any claims not supported by the text?\\
\phantom{xx}Did you fabricate cultural or historical details?\\
\phantom{xx}Did you over-interpret ambiguous evidence?\\
2. Confidence Calibration: Were confidence scores accurate given evidence strength?\\
3. Reasoning Quality: Did you cite specific textual evidence for all claims?\\
4. Layer Boundary Check: Did you accidentally evaluate emotions (L5)\\
\phantom{xx}or philosophical themes (L6)?
\end{quote}

\textit{Layer~6 (Existential)}: Round~2 addresses \textit{Western framework imposition}, a risk unique to philosophical evaluation. The prompt explicitly names non-Western traditions the evaluator must not overlook:

\begin{quote}
\small\ttfamily
ROUND 2/5: PROJECTION BIAS CHECK

1. Western Existentialism Imposition: Did I project Sartre/Camus/Heidegger\\
\phantom{xx}frameworks onto a text from a different philosophical tradition?\\
\phantom{xx}Did I overlook Eastern depth (Buddhist impermanence, Confucian\\
\phantom{xx}relational ethics, Daoist acceptance)?\\
2. Explicit Philosophy Bias: Did I equate explicit philosophical discourse\\
\phantom{xx}with depth, penalizing themes expressed through action or structure?\\
3. Profundity Bias: Did I assume dark or tragic themes are more existentially\\
\phantom{xx}profound than themes of joy, connection, or ordinary life?\\
4. Implicit Existentialism Omission: Did I miss existential depth expressed\\
\phantom{xx}in non-Western vocabulary (Buddhist anicca/dukkha/anatta, Confucian\\
\phantom{xx}li/ren/he, Daoist wu-wei)?
\end{quote}

\medskip
\noindent\textbf{Independent Validator Prompt (condensed).}
The validator receives the complete five-round iterative conversation and applies the following system prompt for single-pass cross-verification:

\begin{quote}
\small\ttfamily
You are an INDEPENDENT VALIDATOR for literary analysis.

YOUR RESPONSIBILITIES:\\
1. Projection Bias Detection: Identify if the iterative evaluator projected\\
\phantom{xx}assumptions about emotional or philosophical expression; flag cultural\\
\phantom{xx}bias in applied norms.\\
2. Hallucination Detection: Flag claims not grounded in the text; identify\\
\phantom{xx}over-interpretations; check for layer boundary violations.\\
3. Reasoning Quality Assessment: Evaluate coherence of arguments; check\\
\phantom{xx}whether evidence supports interpretations.\\
4. Confidence Calibration: Flag overconfidence (high confidence with ambiguous\\
\phantom{xx}content) and underconfidence (low confidence with clear markers).\\
5. Independent Scoring: Provide own dimension scores based on the text;\\
\phantom{xx}compare with iterative evaluator's scores; explain agreements and\\
\phantom{xx}disagreements with specific textual evidence.

CRITICAL MINDSET: Be skeptical but fair. Do not automatically agree.\\
Focus on evidence quality, not score numbers. Avoid your own\\
projection bias in the process of detecting the evaluator's.
\end{quote}

The complete prompt templates and JSON output schemas for all three layers are available at \url{https://github.com/tisage/sage-eval}.

\end{document}